\documentclass[letterpaper]{article} 
\usepackage[draft]{aaai2026}  
\usepackage{times}  
\usepackage{helvet}  
\usepackage{courier}  
\usepackage[hyphens]{url}  
\usepackage{graphicx} 
\usepackage{natbib}  
\usepackage{caption} 
\usepackage{comment}
\usepackage{algorithm}
\usepackage{algpseudocode}
\usepackage{amsmath, amssymb}
\usepackage{graphicx}
\usepackage{bm}
\usepackage{algpseudocode}
\usepackage{amsmath}
\usepackage{caption}
\usepackage{lipsum}
\usepackage{pifont}      
\usepackage{booktabs}    
\usepackage{graphicx}
\usepackage{subcaption}
\usepackage[table,xcdraw,dvipsnames]{xcolor}  
\usepackage{pifont} 
\usepackage{listings}
\usepackage{tcolorbox}
\tcbuselibrary{listings, skins}
\definecolor{pybggray}{RGB}{245,245,245}     
\definecolor{pykeyword}{RGB}{0,128,0}        
\definecolor{pystring}{RGB}{34,139,34}       
\definecolor{pycomment}{RGB}{41,171,135}     
\definecolor{pynumber}{RGB}{0,128,0}      
\definecolor{pyoperator}{RGB}{138,43,226}    
\definecolor{pyfunction}{RGB}{0,0,200}       
\definecolor{MidBlue}{RGB}{70,130,200}  

\lstdefinestyle{pythonstyle}{
  language=Python,
  backgroundcolor=\color{pybggray},
  basicstyle=\ttfamily\small,
  keywordstyle=\color{pykeyword}\bfseries,
  stringstyle=\color{pystring},
  commentstyle=\color{pycomment}\itshape,
  identifierstyle=\color{black},
  emphstyle=\color{pyfunction}\bfseries,               
  emph={function_approximate, graph_approximate, predict, update}, 
  morekeywords={as, with, lambda, yield, from, nonlocal, global, async, await},
  showstringspaces=false,
  breaklines=true,
  tabsize=2,
  frame=none,
  numbers=none,
  xleftmargin=0pt,
  columns=fullflexible,
  aboveskip=0pt,
  belowskip=0pt,
  literate=
    *{0}{{{\color{pynumber}0}}}{1}
     {1}{{{\color{pynumber}1}}}{1}
     {2}{{{\color{pynumber}2}}}{1}
     {3}{{{\color{pynumber}3}}}{1}
     {4}{{{\color{pynumber}4}}}{1}
     {5}{{{\color{pynumber}5}}}{1}
     {6}{{{\color{pynumber}6}}}{1}
     {7}{{{\color{pynumber}7}}}{1}
     {8}{{{\color{pynumber}8}}}{1}
     {9}{{{\color{pynumber}9}}}{1}
     {+}{{{\color{pyoperator}+}}}{1}
     {*}{{{\color{pyoperator}*}}}{1}
     {@}{{{\color{pyoperator}@}}}{1}
     {-}{{{\color{pyoperator}-}}}{1}
     {/}{{{\color{pyoperator}/}}}{1}
     {=}{{{\color{pyoperator}=}}}{1}
}

\newtcolorbox{pythonbox}[1][]{%
  listing only,
  listing engine=listings,
  colback=pybggray,     
  colframe=pybggray,    
  boxrule=0pt,          
  left=0pt,
  right=0pt,
  top=0pt,
  bottom=0pt,
  boxsep=0pt,           
  enhanced,
  sharp corners,
  frame hidden,
  overlay={},           
  valign=center,
  halign=left,
  listing style=pythonstyle,
  #1
}

\usepackage{newfloat}
\usepackage{listings}
\DeclareCaptionStyle{ruled}{labelfont=normalfont,labelsep=colon} 

\floatstyle{ruled}
\newfloat{listing}{tb}{lst}{}
\floatname{listing}{Listing}
\title{Data-Driven Discovery of Interpretable Kalman Filter Variants\\ through Large Language Models and Genetic Programming}
\author {
    Vasileios Saketos\textsuperscript{\rm 1, \rm 2},
    Sebastian Kaltenbach\textsuperscript{\rm 1},
    Sergey Litvinov\textsuperscript{\rm 1},
    Petros Koumoutsakos \textsuperscript{\rm 1, \rm *}
}
\affiliations {
    \textsuperscript{\rm 1}Computational Science and Engineering Laboratory, Harvard University, Cambridge, USA\\
    \textsuperscript{\rm 2}KTH Stockholm\\
    \textsuperscript{\rm *} Corresponding author: petros@seas.harvard.edu
}

\begin{document}

\maketitle

\begin{abstract}
Algorithmic discovery has traditionally relied on human ingenuity and extensive experimentation. Here we investigate whether a prominent scientific computing algorithm, the Kalman Filter, can be discovered through an automated, data-driven, evolutionary process that relies on
Cartesian Genetic Programming (CGP) and Large Language Models (LLM).  We evaluate the contributions of both modalities (CGP and LLM)  in discovering the Kalman filter under varying conditions. Our results demonstrate that our framework of CGP and LLM-assisted
evolution converges to near-optimal solutions when Kalman optimality
assumptions hold. When these assumptions are violated, our framework
evolves interpretable alternatives that outperform the Kalman filter.
These results demonstrate that combining evolutionary algorithms and
generative models for interpretable, data-driven synthesis of simple computational modules is a potent approach for algorithmic discovery in scientific computing.
\end{abstract}


\section{Introduction}

In the past decades, algorithm design has been a manual process that relies heavily on domain knowledge and advanced mathematics.  Recent advances in machine learning, particularly in program synthesis via Large Language Models \cite{RomeraParedes2024,novikov2025alphaevolve,surina2025algorithm} as well as evolutionary computation \cite{Cortacero2023}, now open the possibility of automating parts of this process. A key area of interest is the development of methods that generate algorithms that are interpretable and can adapt to complex real-world environments.\\
This work explores algorithmic discovery in the context of estimating unknown variables based on time-series data, using the Kalman filter~\cite{kalman1960new}, a prominent scientific computing algorithm, as a canonical target. The Kalman filter is a classical, recursive algorithm for estimating the state of a linear dynamical system observed through noisy measurements. Under certain assumptions (i.e. a linear system and Gaussian noise), it yields the optimal minimum mean squared error (MSE) estimator. Various variants such as the Extended Kalman Filter (EKF)~\cite{gelb1974applied} and the Unscented Kalman Filter (UKF)~\cite{Julier1997NewEO} exist that allow the aferomentioned assumptions to be relaxed. While these variants preserve the recursive nature of the original algorithm, they are not optimal estimators and can lead to a degrade in performance.\\
In parallel, deep learning models such as Recurrent Neural Networks (RNNs)~\cite{rumelhart1986learning}, Long Short-Term Memory (LSTM) networks~\cite{hochreiter1997long}, and Gated Recurrent Units (GRUs)~\cite{cho2014learning} have been proposed to estimate unknown quantities using a learned relation based on given data. Despite their success in capturing nonlinear behaviors, such models lack interpretability, require large training sets, and often generalize poorly outside their training distribution.\\
This paper proposes a data-driven framework for discovering the state of a system based on noisy observations. Our framework combines the adaptability of data-driven methods with the interpretability and efficiency of classical solutions. The approach relies on modern tools —specifically Cartesian Genetic Programming (CGP) and Large Language Models (LLMs) — to search over symbolic representations of algorithms using only input–output trajectories and a black-box fitness function. Our primary goal is not to rediscover the Kalman filter structure but to discover new data-driven variants that can compute accurate estimators for scenarios where it was previously impossible.\\

The key contributions of this work are as follows:

\begin{itemize}
    \item We introduce an algorithm discovery framework based on Cartesian Genetic Programming and LLM-assisted evolutionary search (ES).
    \item We evaluate both the LLM-assisted ES and CGP for algorithmic discovery.
    \item We evaluate our framework under a range of challenging scenarios, including non-linearity, non-Gaussian noise, and irregular time sampling — and show that it produces suitable algorithms for these settings.
    \item We release all code and discovered algorithms. The code is available here \footnote{URL will be added upon publication}. 
\end{itemize}

\section{Related Work}
\textbf{Genetic Programming (GP)} is a framework within evolutionary computation that evolves computer programs by simulating principles of natural evolution, including selection, mutation, and crossover \cite{koza1992genetic, 10.5555/1796422}. Cartesian Genetic Programming (CGP) \cite{10.1145/2739482.2756571} extends GP by introducing a structured representation of the solution space in the form of fixed-length directed acyclic graphs, thereby facilitating more localized and stable mutations. 
In general, CGP algorithms maintain a population of candidate solutions that are iteratively refined based on random sampling and mutations as well as the performance on a pre-defined fitness function. A central challenge in CGP lies in the design of mutation operators and the choice of building blocks (the graph nodes), which critically affect the efficiency and effectiveness of the search process. The selection of the building blocks and mutation operators is inherently empirical, and suboptimal choices can significantly degrade the performance of the evolved programs. Since these elements are typically problem-specific, their design requires knowledge and intuition about the nature of the problem. Recently, CGP has been successfully adopted to discover algorithms for biomedical image analysis \cite{Cortacero2023} using primitive image modification operators as building blocks. For the present work, we choose building blocks based on fundamental mathematical operations to allow for general algorithmic discovery.\\
\textbf{LLMs} have not only  revolutionized the field of Natural Language Processing, but have been successfully applied  across numerous scientific domains. Their success would have not been possible without the advent of the Transformer architecture~\cite{vaswani2017attention} that allowed the development of earlier models such as BERT \cite{devlin2019bert} and GPT \cite{brown2020language}. Recently, reasoning capabilities have become a central focus in LLM development, with DeepSeek~\cite{guo2025deepseek} emerging as first open-source model to rival the performance of closed-source alternatives \cite{comanici2025gemini} by combining large-scale pretraining with reinforcement learning techniques to enhance logical inference capabilities. As such LLMs in general consist of billions of parameters, techniques such as QLoRA~\cite{dettmers2023qlora} have been developed to enable memory-efficient inference and fine-tuning through 4-bit quantization and parameter-efficient adaptation.\\
Automatic algorithmic discovery can be based on LLMs as they can be employed to generate new program variants by using their ability to produce a wide spectrum of syntactically and semantically meaningful mutations based on representations learned during pretraining, thereby eliminating the need for manually defined primitive sets or mutation operators as in GP. Recently, Funsearch~\cite{RomeraParedes2024} was introduced as a hybrid framework that combines Large Language Models with evolutionary search. It enables the automated generation of high-quality program variants guided by task-specific evaluation. Through the use of Funsearch, the authors were able to automatically generate state-of-the-art heuristics for several challenging combinatorial problems, including bin packing \cite{10.5555/2183}, the cap set \cite{capset}, and the admissible sets problem \cite{Tao_Vu_2006}. AlphaEvolve~\cite{deepmind2025alphaevolve} is a recently introduced coding agent that extends the capabilities of Funsearch by enabling the evolution of entire source files rather than isolated functions. This led to the discovery of state-of-the-art algorithms, including a novel method for \(4 \times 4\) complex matrix multiplication that requires only 48 scalar multiplications~\cite{strassen1969gaussian, deepmind2025alphaevolve}. Our framework relies both on CGP and LLMs and the LLM part is heavily inspired by Funsearch~\cite{RomeraParedes2024} and its successor  AlphaEvolve~\cite{deepmind2025alphaevolve}  but as no open-source software has been released for both of these framework, we implemented our LLM-assisted evolutionary search framework ourselves.

\section{The Kalman Filter}
The Kalman filter is an important algorithm in scientific computing used for estimating the internal state of a dynamical systems from (potentially noisy) observations.
Originally introduced by Rudolf K\'{a}lm\'{a}n \cite{kalman1960new} for systems with discrete time steps and later extended for continuous time systems \cite{kalman1961new}. Its effectiveness in early aerospace applications, such as the Apollo navigation system \cite{1977fmds.book.....B}, highlighted its practical value and efficiency. In contemporary cyber-physical systems, Kalman filtering continues to serve as a fundamental tool in domains such as robotics, autonomous navigation, and sensor fusion \cite{grewal2015kalman}.\\
The discrete-time Kalman filter is designed for systems modeled as linear dynamical processes disturbed by Gaussian noise. It assumes that the system dynamics and observation models are linear, with both processes subject to Gaussian noise. Let \( x_t \in \mathbb{R}^n \) denote the hidden state of the system at time \( t \), and \( z_t \in \mathbb{R}^m \) the observed measurement. The system evolves according to the following discrete-time linear state-space equations:
\begin{equation}
x_t = F x_{t-1} + B u_{t-1} + w_{t-1}, \quad w_{t-1} \sim \mathcal{N}(0, Q)
\end{equation}
\begin{equation}
z_t = H x_t + v_t, \quad v_t \sim \mathcal{N}(0, R)
\end{equation}

Where, \( F \in \mathbb{R}^{n \times n} \) is the state transition matrix, \( B \in \mathbb{R}^{n \times k} \) is the control input matrix, \( H \in \mathbb{R}^{m \times n} \) is the observation matrix. \( Q \) and \( R \) are the covariance matrices of the process and observation noise respectively. The control input \( u_{t-1} \) is assumed to be known.

The Kalman filter operates recursively in two main steps: predict and update. In the predict step, it projects the current state and its uncertainty forward in time, estimating the system's next state without incorporating new observations.
\begin{equation}
\hat{x}_{t|t-1} = F \hat{x}_{t-1|t-1} + B u_{t-1}
\label{eq:1}
\end{equation}
\begin{equation}
P_{t|t-1} = F P_{t-1|t-1} F^\top + Q
\label{eq:2}
\end{equation}

Where, \( \hat{x}_{t|t-1} \) is the prior estimate of the state at time \( t \), and \( P_{t|t-1} \) is the corresponding error covariance matrix.

Upon receiving the observation \( y_t \), the update step adjusts the prediction using the measurement:

\begin{equation}
K_t = P_{t|t-1} H^\top \left( H P_{t|t-1} H^\top + R \right)^{-1}
\end{equation}
\begin{equation}
\hat{x}_{t|t} = \hat{x}_{t|t-1} + K_t (y_t - H \hat{x}_{t|t-1})
\end{equation}
\begin{equation}
P_{t|t} = (I - K_t H) P_{t|t-1}
\end{equation}

The matrix \( K_t \) is the Kalman gain, determining the relative weighting between the predicted state and the new observation. The posterior estimate \( \hat{x}_{t|t} \) combines the prior with the measurement residual, while \( P_{t|t} \) reflects the reduced uncertainty after the update.\\
The Kalman filter is optimal under the assumptions of linearity and Gaussian noise, providing the minimum mean squared error (MMSE) estimate of the state. However, if the noise deviates from Gaussianity, or the system dynamics are nonlinear, the theoretical guaranties do not longer hold and the filter's performance can degrade. In such cases, extensions like the Extended Kalman Filter (EKF) or Unscented Kalman Filter (UKF) are employed. Nevertheless, these methods are not theoretically optimal and often require additional information, such as the computation of Jacobians in the case of EKF, or careful parameter tuning and prior knowledge of system structure, which may limit their applicability in complex or poorly understood systems.

\section{Methodology}
\label{sec:frameworks}
Within this work, we phrase algorithmic discovery as an optimization problem which is subsequently solved using an iterative refinement procedure. We do not need to provide any details regarding the actual algorithm that is going to be discovered, but only an evaluation function that can compute a score for each suggested algorithm.
\begin{figure*}[h!]
\centering
\includegraphics[width=1.0\textwidth]{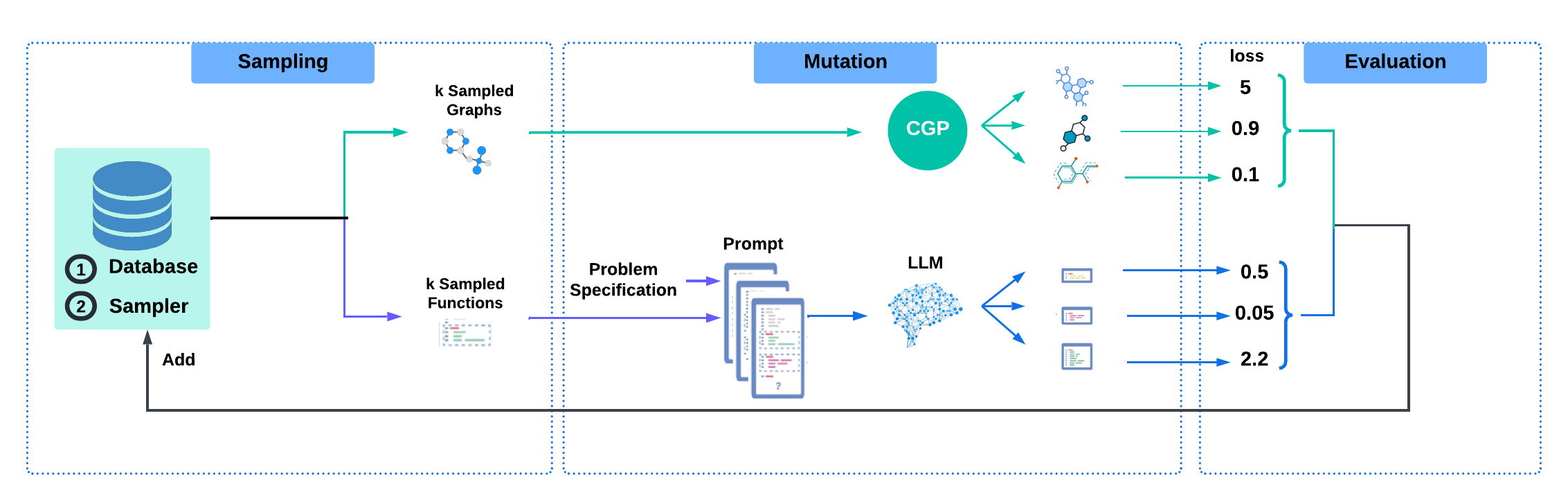} 
\vspace{-1.em}

\caption{General overview of our framework for algorithmic discovery: Algorithm candidates are stored in a database. To generate new algorithm variants using either CGP or LLM-assisted ES, algorithms are samples from the database and mutated. Subsequently, the new algorithm variants are evaluated using a fitness function and newly identified top performing algorithms are incorporated into the database.}
\label{fig:processdemonstration}
\end{figure*}
Our framework builds both on LLM as well as Cartesian Genetic Programming. A general overview can be found in Figure \ref{fig:processdemonstration}.\\
At the core of the framework is a database that contains the current best performing algorithms discovered by both LLM and CGP and their corresponding scores.  Despite operating in distinct
solution spaces, both the CGP and LLM based approach aim to solve the same optimization problem. As CGP is operated on CPUs and the LLMS on GPU this process can be easily parallelized and we can efficiently use all available hardware resources.
The framework operates using three phases: Sampling,
Mutation, and Evaluation. In the sampling stage, solutions are sampled from the database, favoring those with higher fitness value. During the mutation phase, both the CGP and the LLM based approach apply transformations to
the sampled candidates with the objective of generating improved
solutions. In Cartesian Genetic Programming (CGP), mutations involve
altering node connections or changing the type of computational nodes,
which are chosen from a predefined set of functions. The graph
structure has a fixed size, and mutations operate within these
constraints to explore different computational pathways. In contrast,
for the LLM-assisted evolutionary search, mutations are applied by modifying the structure of symbolic
functions, allowing for more flexible and expressive
changes. This approach can dynamically introduce new syntactic elements,
effectively expanding the representation space as the search
progresses. As shown in Figure~\ref{fig:processdemonstration},
the system takes a problem specification along with two sampled
candidate solutions and creates the prompt.  The prompt is forwarded
to the actual LLM, which in our case is a publicly available model from DeepSeek \cite{guo2025deepseek}. The LLM processes the prompt and produces mutations and
combinations of the two input functions.\\
After generation, each candidate solution is evaluated based on a fitness/loss function. Although the CGP- and LLM-assisted approaches employ different internal
representations, they are both designed to operate over the same input
and output interfaces. This alignment enables the use of a single,
well-defined lfitness function to assess solution quality across both
approaches in parallel. The
functions and fitness values are used to update the database that maintains
the top {\it N} highest-scoring candidates.\\
A more detailed description of both approaches can be found below:

\subsection{LLM assisted evolutionary search}
Our LLM-assisted ES is strongly inspired by Funsearch \cite{RomeraParedes2024}. However, as no official publicly available code is available, we implemented it ourselves based on the available descriptions. This implementation was done utilizing the transformers library of Hugging Face \cite{wolf2019huggingface} and we employed the 
DeepSeek-R1-Distill-Qwen-14B model \cite{guo2025deepseek}. We note that we also tested other models within the DeepSeek family but models with fewer than 14 billion parameters tended to produce repetitive and low-quality generations, whereas larger models introduced significant latency and substantially increased computational costs. We thus identified the 14B model as offering the best trade-off between efficiency and output quality. To ensure that the obtained framework is comparable to Funsearch, we applied it to the backpacking task as defined in the Funsearch paper \cite{RomeraParedes2024} and obtained comparable results. A detailed description of this experiment and its results can be found in Appendix A. \\

For all our experiments, we employed a solution database that retains the 200 best solutions and their corresponding fitness value.
For each mutation step,  60 candidates are sampled.  The selection probability \( P_i \) for the \( i \)-th candidate is given by:
\vspace{-0.5em}
\[
P_i = \frac{e^{-\frac{f_i}{T}}}{\sum_{j} e^{-\frac{f_j}{T}}}
\]
where \( f_i \) denotes the fitness value of the \( i \)-th candidate, and \( T  \) is the temperature parameter controlling the trade-off between exploitation and exploration. This mechanism favors higher-quality solutions while still allowing exploration based on including lower-performing ones. Afterwards, we instruct the LLM to generate combinations and mutations, where each prompt has as an input two of the chosen candidate functions from the data base as well as a prototype that the generated functions must adhere to. A key component of our approach is to generate multiple variants (mutations) for each prompt. This capability was largely impractical with earlier generation language models such as PaLM \cite{chowdhery2023palm} and StarCoder \cite{li2023starcoder}, which more frequently produced repetitive or poorly aligned outputs that did not adhere to the specified function definitions.
We set the number of output tokens of the LLM $max\_len$ = 3,000 tokens as a trade-off between computational efficiency and exploration.\\
We note, that the ES is distributed across four independent islands as suggested in \cite{RomeraParedes2024} and each island is executed on a separate GPU and uses its own database. After 10 iterations of sampling, mutation and updating the database, the stored algorithms of the islands with the weakest top-performing candidates are cleared and reinitialized with the best candidate from the island that achieved the highest performance.

\subsection{Cartesian Genetic Programming}
Our implementation of CGP is inspired by the work of Cortacero et al.~\cite{Cortacero2023}. The solutions are represented as directed acyclic graphs with a fixed maximum size. In each mutation step, we choose 30 graphs from the database using the same sampling rule as in the LLM-assisted ES. For each chosen graph, we generate 1000 variants. The mutations involve altering node connections or changing the type of computational nodes, which are chosen from a predefined set of functions. In our case this set of functions include addition (+), assignment (=), matrix multiplication (@), and matrix inversion (inv). In our CGP setup, we employ the island model exactly as used in the LLM-assisted ES with the difference that each island uses a pre-defined amount of CPU resources instead of a GPU.\\
The hyperparameters used for both approaches are summarized in Appendix B.

\begin{table*}[]
\centering
\setlength{\tabcolsep}{28pt}
\renewcommand{\arraystretch}{1.1}
\begin{tabular}{lccc}
\toprule
\textbf{Method} & \textbf{CGP} & \textbf{LLM-assisted ES} & \textbf{Random Search} \\
\midrule
\textbf{predict}               
& \cellcolor{green!60}$0.995077  \pm 9\mathrm{e}{-3}$
& \cellcolor{green!45}$0.995493  \pm 9\mathrm{e}{-3}$
& \cellcolor{yellow!30}$1.009967  \pm 9\mathrm{e}{-3}$ \\

\textbf{predict + update}      
& \cellcolor{green!45}$0.995220 \pm 9\mathrm{e}{-3}$ 
& \cellcolor{red!25}$1.968045 \pm 1\mathrm{e}{-2}$
& \cellcolor{red!15}$1.209344 \pm 1\mathrm{e}{-2}$\\
\bottomrule
\end{tabular}
\caption{MSE loss ± standard error of our methods compared to Kalman Filter performance (0.995077). 
\textcolor{green!60!black}{Green} highlights optimal or near-optimal, 
\textcolor{orange!80!black}{yellow/orange} indicates moderate deviation, and 
\textcolor{red!80!black}{red} marks the least favorable outcomes in each row.}
\label{tab:method_comparison}
\end{table*}

\section{Experiments}
We apply the proposed framework for scientific discovery to datasets from dynamical systems and try to find an algorithm that predicts the state representation based on noisy observations. While we know that for linear dynamical systems with Gaussian noise the resulting algorithm should be the Kalman Filter due to its theoretical optimality, for other dynamical systems of interest with different added noise the optimal estimation algorithm is in general unknown.\\
To generate the trajectory data, we consider a linear time-invariant, discrete-time dynamical system that is modeling the position and velocity of an object. This system is governed by the following equation \(
{x}_k = {F} {x}_{k-1} + {G} a_k \), where
the state \( {x}_k = [p_k,\, v_k]^T \) includes position and
velocity, and the random acceleration input \( a_k \sim \mathcal{N}(0,
\sigma_a^2) \) induces process noise \( {w}_k = {G} a_k
\) with covariance \( Q = {G} {G}^\top
\sigma_a^2 \). F describes the system's dynamics. The system is fully observable via a linear
measurement model \( {z}_k = {H} {x}_k +
{v}_k \), with \( {H} = {I} \) and measurement
noise \( {v}_k \sim \mathcal{N}({0}, \sigma_z^2
{I}) \). More details can be found in the Appendix C.\\
The described system serves as a canonical benchmark for filtering and control
tasks~\cite{greenberg2023optimization,freirich2023perceptual} and in the present form fulfills all constraints such that Kalman-Filer is the optimal estimator for the state based on noisy observations.\\
We note that each algorithm candidate takes as input the current state estimate, system dynamics, covariance matrices, and the latest observation. 
To assess performance and compute the fitness value of each algorithm candidate, we compute the mean squared error (MSE) between
the estimated $\hat{{x}}$ and true states $x$ on a trajectory of length \( T \): \begin{equation}
\mathcal{L} = \frac{1}{T} \sum_{k=1}^T \|\hat{{x}}_k -
{x}_k\|^2 
\end{equation}
As evaluating the performance of candidates for algorithms is done multiple times, we only use a single trajectory of 200 time steps for these evaluations during training to ensure rapid evaluation and high-throughput during the discovery process.\\

To ensure robustness and generalization, all candidates are evaluated
on a substantially larger validation set comprising \( 50 \)
independently simulated trajectories, each of length \( 500 \). The
best-performing algorithm on this set is selected for final evaluation
on an unseen test set of equal size and structure. This protocol
ensures that final performance reflects true generalization to new
trajectories and noisy realizations, rather than overfitting to a
specific training instance.
For each of our experiments, we ran the LLM-assisted ES twice on four H100 GPUs each whereas the CGP was executed a total of fifteen times using sixty-four CPUs each. Each run
was executed over a period of a maximum of 72 hours.\\
To establish a baseline for our algorithm's performance, we use random
search. Random search generates K graph nodes and connects them
arbitrarily to form valid computational graphs. The connections ensure
a valid path from inputs to outputs, without any optimization or
heuristics. 

\subsection{(Re)discovering the Kalman Filter}

For the first experiment, we choose the dynamical system as described in the previous section which fulfills all the criteria for the optimality of the Kalman Filter. 
We note that apart from the evaluation function that is used to score each discovered algorithm, no further details about the Kalman Filter are provided and we are uncovering the structure and behavior of the Kalman filter purely data-driven through finding the best fitness value — specifically, by minimizing the mean squared error (MSE) between its state estimates and the true latent states over time. \\
We evaluate performance on two principal benchmarks. The first, and conceptually simplest, assesses the ability to reconstruct the Kalman filter's prediction step, assuming that the update step is already known. This corresponds to equation \ref{eq:1} and \ref{eq:2} in the original Kalman-Filter that we attempt to rediscover. The second benchmark requires reconstructing the full Kalman filter, including both the prediction and update steps.\\

In case of the task to to rediscover the first half of the Kalman filter only, i.e. the predict operator, both CGP and LLM-assisted ES receive the state transition matrix \( F \), current state \( x \), covariance \( P \), and process noise \( Q \) as an input, and must output the predicted state \( x_{\text{predict}} \) and updated covariance \( P \). In the full Kalman filter task, they are also given the measurement \( z \) and measurement noise \( R \), and must produce \( x_{\text{predict}}, P, y \) (innovation), \( S \) (innovation covariance), \( K \) (Kalman gain), and \( x_{\text{update}} \) (updated state).\\
However, we note that we pass all inputs without any description and even use a generic representation. Together with not providing our framework any information regarding the semantic nature of the task or the internal structure of the target algorithm, this is done to rigorously prevent data leakage. Especially in case of the LLM-assisted ES, it can be assumed that the LLM has seen the Kalman-Filter during training and including a task description or variable names in the prompt would significantly accelerate the discovery. Instead, we only explicitly define the target function signature, specifying the exact number and order of input and output variables using a generic names (i.e. inputs $i_1, \ldots, i_n$; outputs $o_1, \ldots, o_m$). This guarantees that the search process is limited to a fixed function design, but prevents passing any indirect hints when generating programs. We note that data leakage is not a concern for the CGP approach.\\

Table \ref{tab:method_comparison} presents the MSE loss and the standard error for different methods, benchmarked against the Kalman filter, which serves as the lower bound. We note that CGP successfully discovered the optimal \textit{predict} program, matching the Kalman filter’s performance, while LLM-assisted ES closely approximated it. As CGP operates within a bounded search space, i.e. a finite number of nodes and building blocks, the complexity of the solution is constrained although the actual number of possible combination of these building blocks can still be very high. For LLM-assisted ES, by contrast, very long and complicated candidate algorithms can be generated that tend to overfit which negatively impacts the evolutionary search close to the optimal solution.\\
For the full predict + update task, CGP maintains near-optimal performance. In contrast, our LLM-assisted ES tends to appear to stuck in local minima. This limitation is likely due to computational constraints: Firstly, the utilized LLM is using 14 billion parameters only and larger models could lead to better performance. Secondly, we limit the number of generated tokens per prompt, which reduces the number of new candidates that are generated in the mutation steps. The difference in performance between CGP and LLM-assisted ES further justifies our approach of employing both methods to establish a robust framework for scientific discovery across diverse settings and tasks. We note that in Appendix D, we present an ablation study where we progressively increase the difficulty by increasing the percentage of the Kalman filter that is to be discovered. There, we analyze in detail when the drop in performance of the LLM-assisted ES occurs.
Random Search discovered a competitive program for the simple \textit{predict} task but degraded for the full Kalman Filter. This is expected, as more complex problems yield a significantly larger solution space, reducing the probability of discovering optimal solutions through purely random exploration. Our framework significantly outperforms this baseline for both tasks.

\subsection{Beyond the Kalman filter}

\begin{figure*}[t]
\centering
\vspace{-0.5em}

\caption*{Legend: 
\textcolor{MidBlue}{\textbf{- - -}} Observations, 
\textcolor{Orange}{\textbf{···}} Kalman filter, 
\textcolor{ForestGreen}{\textbf{- - -}} Random Search, 
\textcolor{Red}{\textbf{– · –}} LLM-assisted ES, 
\textcolor{RoyalPurple}{\textbf{—}} CGP.
}
\vspace{-1.em}
\begin{subfigure}[t]{0.33\textwidth}
    \centering
    \includegraphics[width=\linewidth]{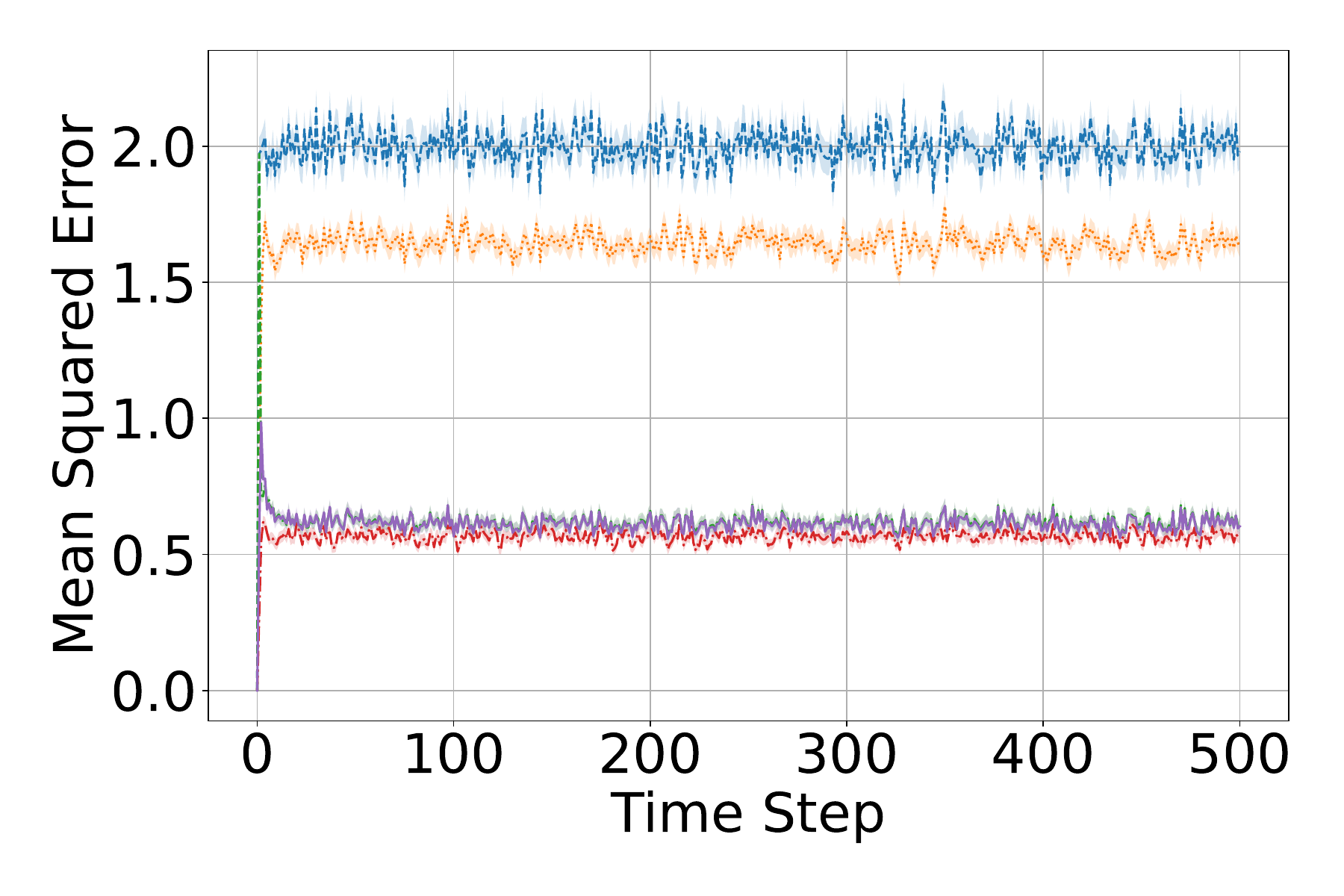}
    \caption{Half Gaussian Noise}
    \label{fig:plot1}
\end{subfigure}
\hfill
\begin{subfigure}[t]{0.33\textwidth}
    \centering
    \includegraphics[width=\linewidth]{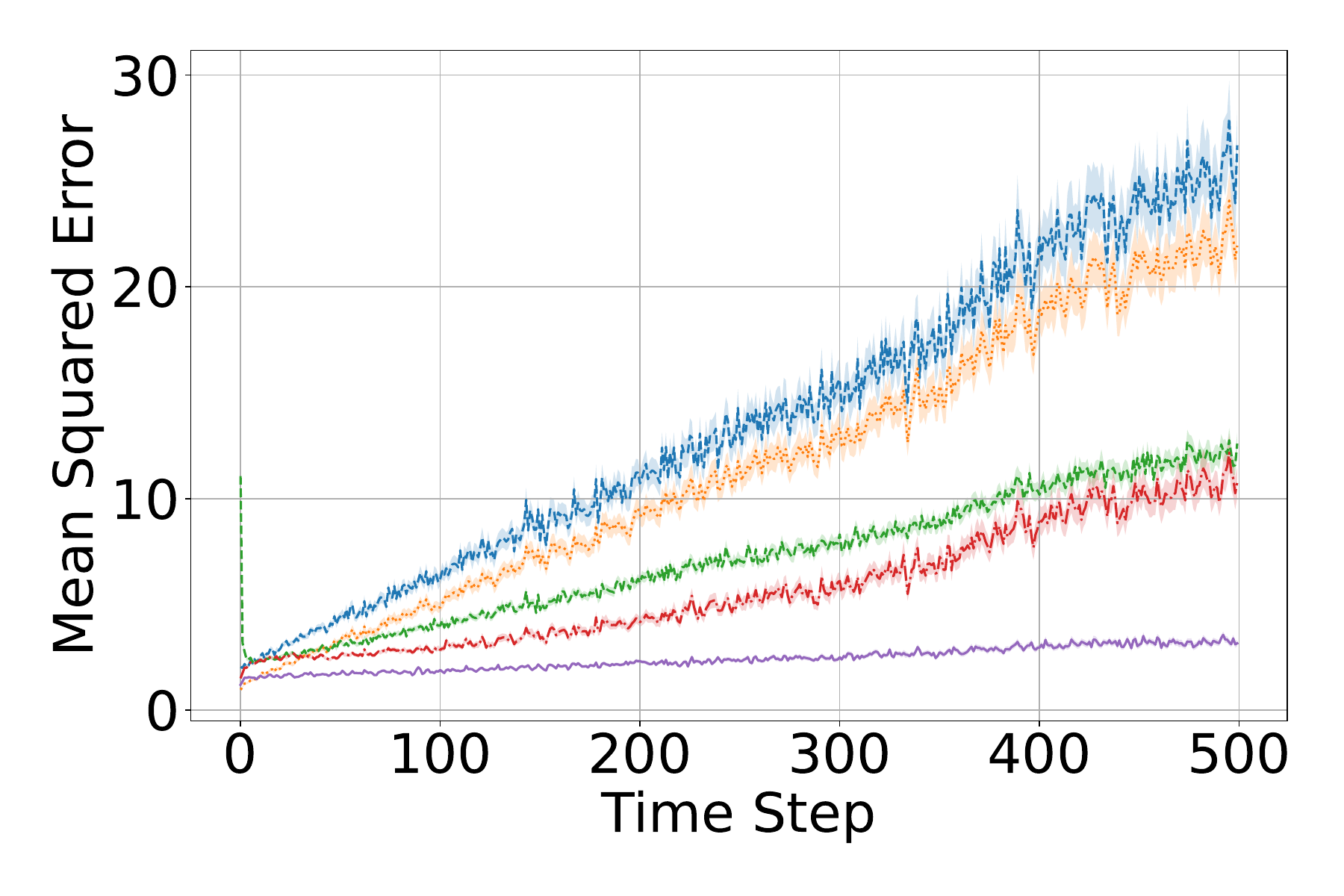}
    \caption{Delayed observations}
    \label{fig:plot2}
\end{subfigure}
\hfill
\begin{subfigure}[t]{0.33\textwidth}
    \centering
    \includegraphics[width=\linewidth]{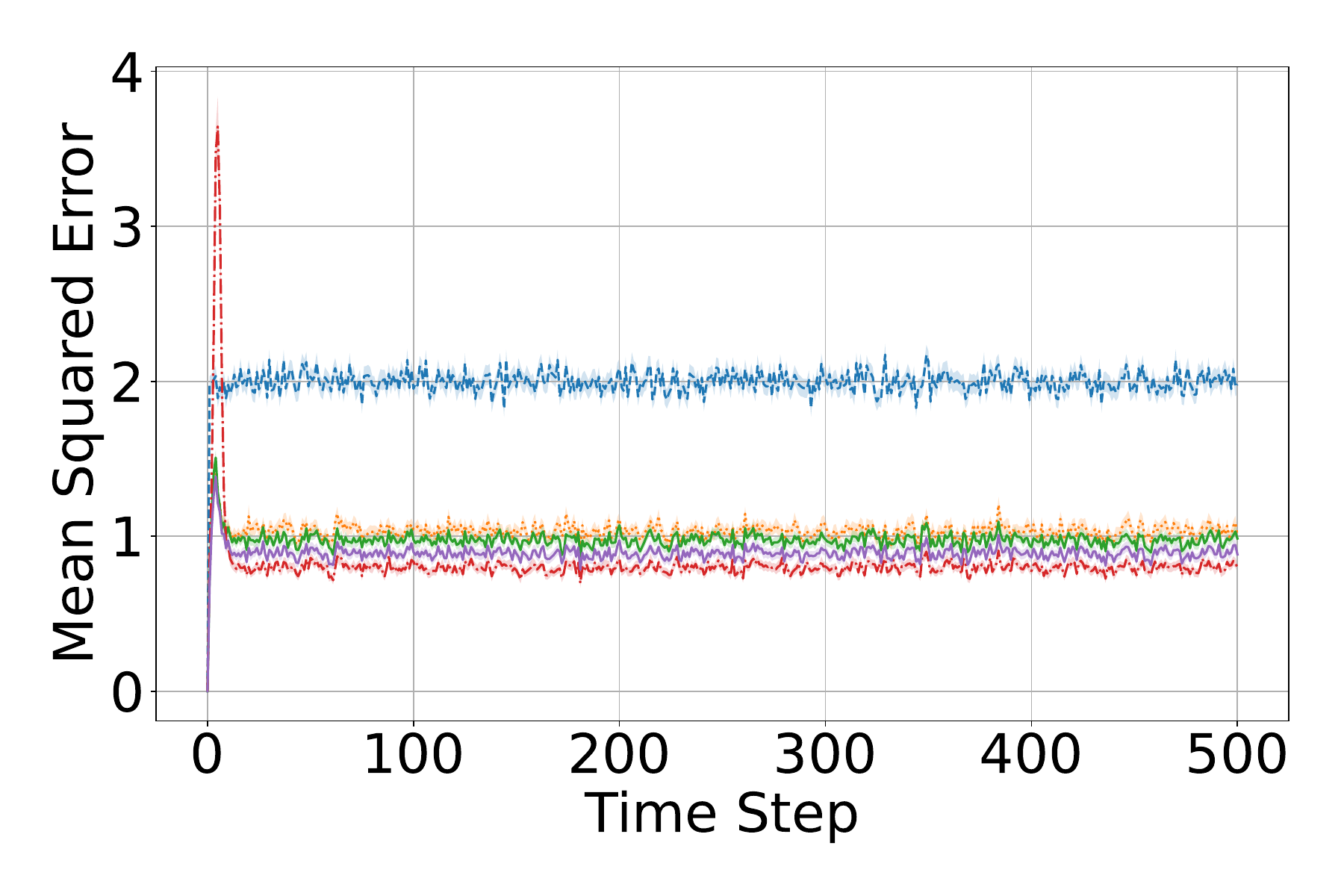}
    \caption{Non-linear dynamics}
    \label{fig:plot3}
\end{subfigure}
\caption{Mean squared error (MSE) of discovered algorithms and baselines for different settings.} 
\label{fig:three_plots} 
\end{figure*}
\begin{table*}[t]
\centering
\renewcommand{\arraystretch}{1.1}
\setlength{\tabcolsep}{7.5pt}
\begin{tabular}{lccccc}
\toprule
 & \textbf{LLM-assisted ES} & \textbf{CGP} & \textbf{Random Search} & \textbf{Kalman filter} & \textbf{Observations} \\
\midrule
\textbf{Half Gaussian Noise} 
& \cellcolor{green!60}$0.5618 \pm 6\mathrm{e}{-3}$
& \cellcolor{yellow!25}$0.6032 \pm 5\mathrm{e}{-3}$
& \cellcolor{orange!30}$0.6116 \pm 5\mathrm{e}{-3}$
& \cellcolor{red!25}$1.6221 \pm 1\mathrm{e}{-2}$
& \cellcolor{red!30}$1.9680 \pm 1\mathrm{e}{-2}$ \\

\textbf{Delayed observation} 
& \cellcolor{yellow!25}$4.9601 \pm 5\mathrm{e}{-1}$
& \cellcolor{green!60}$2.2788 \pm 1\mathrm{e}{-1}$
& \cellcolor{orange!30}$6.9536 \pm 4\mathrm{e}{-1}$
& \cellcolor{red!25}$10.6176 \pm 1\mathrm{e}{-1}$
& \cellcolor{red!30}$16.7031 \pm 2\mathrm{e}{+1}$ \\

\textbf{Nonlinear Dynamics} 
& \cellcolor{green!60}$0.8064 \pm 8\mathrm{e}{-3}$
& \cellcolor{yellow!25}$0.8760 \pm 6\mathrm{e}{-3}$
& \cellcolor{orange!30}$0.9589 \pm 7\mathrm{e}{-3}$
& \cellcolor{red!25}$1.0235 \pm 2\mathrm{e}{-2}$
& \cellcolor{red!30}$1.9680 \pm 1\mathrm{e}{-2}$ \\
\bottomrule
\end{tabular}
\caption{Mean squared error (MSE) of discovered algorithms and baselines under different conditions (mean ± standard error). 
Color gradients indicate performance per row: \textcolor{green!60!black}{green} for best, 
\textcolor{orange!80!black}{orange/yellow} for moderate, and \textcolor{red!80!black}{red} for worst.}
\label{tab:generalization}
\end{table*}
\begin{figure*}[t]
\centering
\noindent
\begin{minipage}[t]{0.49\textwidth}
  \begin{pythonbox}[height=4.cm]
  \begin{lstlisting}[style=pythonstyle]
def function_approximate(x, F, P, Q, z, R):
    xp = F @ x
    P_new = F @ P @ F.T + Q
    y = z - xp
    S = P_new + R + F.min(axis=1)[:, None] * 1.2
    inv_S = np.linalg.inv(S)
    K = (P_new @ inv_S) * 0.85
    x = xp + K @ y
    P = (P_new - K @ S @ K.T) * 0.95
    return xp, P, y, S, K, x
  \end{lstlisting}
  \end{pythonbox}
\end{minipage}%
\hfill
\begin{minipage}[t]{0.49\textwidth}
  \begin{pythonbox}[height=4.cm]
  \begin{lstlisting}[style=pythonstyle]
def graph_approximate(x, F, P, Q, z, R):
    A = R.T + x
    y = P - A
    S =  (y + F).T @ P + A
    K = np.linalg.inv(S)
    x = R - (A + R.T) @ K
    P = np.linalg.inv(y)
    return xp, P, y, S, K, x
  \end{lstlisting}
  \end{pythonbox}
\end{minipage}
\caption{{ Functions discovered for Half Gaussian Noise by LLM-assisted ES on the left and CGP discovered function on the right.}}
\label{fig:discovered_functions}
\end{figure*}

In this section, we investigate the applicability of our methods in scenarios where the Kalman filter is no longer the optimal solution. Our objective is to assess whether our algorithmic discovery frameworks can synthesize robust and interpretable programs that surpass the Kalman filter's performance under violated assumptions. To this end, we consider challenging settings characterized by non-Gaussian noise, delayed observations, and nonlinear system dynamics.  As currently no optimal algorithm exists for these setting, we do not have to worry about data leakage. Thus, in contrast to the previous setting, the language model is provided with complete information about the underlying problem to leverage the full knowledge acquired during pretraining for discovering robust and interpretable algorithms. Moreover, the starting databases are initialized using the original Kalman filter combined with random initializations. We observed that this lead not only to faster convergence but also to algorithms that can be more easily interpreted due to being closer to the roginal Kalman-Filter. We note that in general LLM-assisted ES was relativly robust regarding the initialization as the LLM was also able to correctly reconstruct the Kalman filter from the prompt alone in case none of the initial algorithm corresponded to a Kalman-Filter. In contrast, CGP showed improved performance, likely due to a well-chosen starting point with the original Kalman-Filter and enhanced exploration thanks to the random initializations.

As a first task, we modified the dynamical system from the previous experiment by introducing asymmetric noise via a Half-Gaussian distribution. We apply our frameworks to assess whether discovered algorithms can outperform the Kalman filter under this violation of classical assumptions.\\
Next, we simulate a setting with randomly delayed observations. Again we consider the dynamcial system from the previous experiment but at each time step, the observation corresponds to a slightly earlier time, with the delay sampled uniformly from a fixed range. Since this delayed time typically falls between two discrete simulation steps, we interpolate between adjacent ground-truth states to generate the corresponding observation. Our model, however, is not informed about this delayed observation and processes this delayed observation as if it were current. This setup mimics real-world sensor latency and breaks the Kalman filter’s assumption of synchronized dynamics and observations, providing a challenging scenario for assessing the robustness of discovered programs.\\ Finally, we evaluate our frameworks in a setting with nonlinear but known dynamics. The state evolves according to \( x_{t+1} = F \cdot g(x_t) + w_t \), where \( g(x) \) introduces structured nonlinearities such as cubic and sinusoidal terms. Although the dynamics deviate from linearity, their structure is assumed to be fully known. The observations remain linear. This experiments tests the ability of our methods to discover effective estimators in settings where classical linear filters, such as the Kalman filter, are no longer applicable.

Figure~\ref{fig:three_plots} presents the MSE and the standard error trajectories over 500 time steps for the three scenarios. The corresponding time-averaged MSE values are reported in Table~\ref{tab:generalization}. In the half-Gaussian noise scenario, the algorithm discovered by LLM-assisted ES achieves a steady-state MSE of 0.56, corresponding to a nearly threefold improvement over the Kalman filter, which plateaus at 1.62. As shown in the leftmost panel, the performance gap between the discovered algorithms and the Kalman filter remains consistently large throughout the time horizon, highlighting the persistent advantage of program-discovered estimators under asymmetric noise conditions.

In the delayed observation setting (Figure~\ref{fig:three_plots}, center), all methods exhibit increasing error over time due to the accumulation of misalignment between the latent state and delayed observations. This effect becomes more pronounced as the object accelerates, making the object’s position increasingly sensitive to temporal discrepancies. The Kalman filter, which assumes perfectly synchronized observations, performs poorly under this violation, yielding a high average MSE of 10.62 (Table~\ref{tab:generalization}). In contrast, the algorithms discovered by LLM-assisted and CGP demonstrate substantially improved robustness, achieving average MSEs of 4.96 and 2.28, respectively. While both methods outperform the baselines, CGP achieves the best performance. This highlights the benefit of relying on two approaches for creating new algorithms to increase the chances of finding a very effective algorithm.

Finally, in the nonlinear setting (Figure~\ref{fig:three_plots}, right), both LLM-assisted ES and CGP achieve lower errors of 0.81 and 0.88 comapred to the Kalman Filter. The learned programs consistently deliver superior accuracy, highlighting their flexibility in adapting to known nonlinear dynamics.

We note that the random search was able to surpass the performance of the Kalman filter but falls short of the algorithms discovered using both LLM-assisted ES and CGP in all three settings.

In Figure~\ref{fig:discovered_functions}, we present the function discovered by LLM-assisted ES (left) and CGP (right) for the half Gaussian noise case.  It is evident that the algorithm discovered by the LLM-assisted ES assigns greater weight to the predicted state \( xp \) compared to the observation. This bias is introduced in multiple ways.
First, the innovation covariance matrix \( S \) is modified by adding the term
\(
F.\min(\text{axis}=1)[:, \text{None}] \times 1.2,
\)
which increases its diagonal elements and thus inflates the perceived uncertainty of the observation. This, in turn, reduces the magnitude of the inverse \( S^{-1} \). Additionally, the Kalman gain \( K \) is explicitly scaled by a factor of \( 0.85 \), further diminishing the influence of the residual \( z - xp \).  Together, these modifications cause the update step to rely more heavily on the prediction \( xp \) than on the observation \( z \), resulting in a conservative correction.

In contrast, the algorithm discovered by CGP is more difficult to interpret as it does not closely resemble the structure of the Kalman filter. This shows that despite consisting only of known building blocks, a detailed analysis would be required to analyze the identified algorithm.

In Appendix E, we present the algorithms discovered for the other two settings together with a short analysis.

\section{Conclusions}
We have presented a framework for algorithmic discovery based on both CGP and LLM-assisted ES. By relying on both CGP and LLM-assisted ES, we are able to discover high-performing algorithms, as our experiments show that either CGP or the LLM yields the best results depending on the setting. Moreover, this approach allows us to simultaneously leverage both GPU and CPU resources. The framework was successfully applied to discover novel Kalman Filter variants and showed promise for discovering fundamental scientific computing algorithms.\\
Current limitations include that there is no exchange of algorithm candidates between the CGP and LLM-assisted ES during the algorithmic discovery process. While a naive exchange of candidates is not possible as both approaches require a specific structure of the algorithms, even just the exchange of a few suitable algorithm candidate towards the end of the algorithmic discovery process could increase exploration significantly in case one of the two approaches is stuck in a local minima.\\  Another limitation is that during the mutation process the LLM may produce very complex patterns that overfit the data. As the complexity of the algorithm of interest increases, it becomes increasingly difficult to trace and understand the rationale behind the LLM's outputs, which may hinder transparency and trust in practical applications. One solution could be to explicitly specify constraints or desired properties in the prompt; however, this approach may not fully prevent the model from generating overly complex or non-intuitive functions, especially when operating in high-dimensional or ambiguous problem spaces.\\
In future work, we plan to develop strategies to exchange algorithm candidates between CGP and LLM-assisted ES. Moreover, we plan to extend this framework to more complex domains, including multi-object tracking in autonomous driving scenarios, as explored in~\cite{chiu2021probabilistic,chiu2024probabilistic}, as well as nonlinear control and decision-making systems~\cite{10.1007/978-3-319-15224-0_1}. Finally, a promising direction is applying our discovery framework to settings where system and control dynamics are unknown, aiming to extract interpretable models purely from observations.

\section*{Acknowledgements}
The authors thank Cengiz Pehlevan and Yue Lu for valuable discussions and helpful insights.\\
S.K. and P.K. acknowledge support by the Defense Advanced Research Projects Agency (DARPA) through Award HR00112490489.\\
V.S. acknowledges support by the Karl Engvers foundation.

\bibliography{aaai2026}

\newpage
\appendix
\appendix
\section{Appendix A: Binpacking}
\label{app:binpack}

\begin{figure}[h!]
    \centering
    \includegraphics[width=0.46\textwidth]{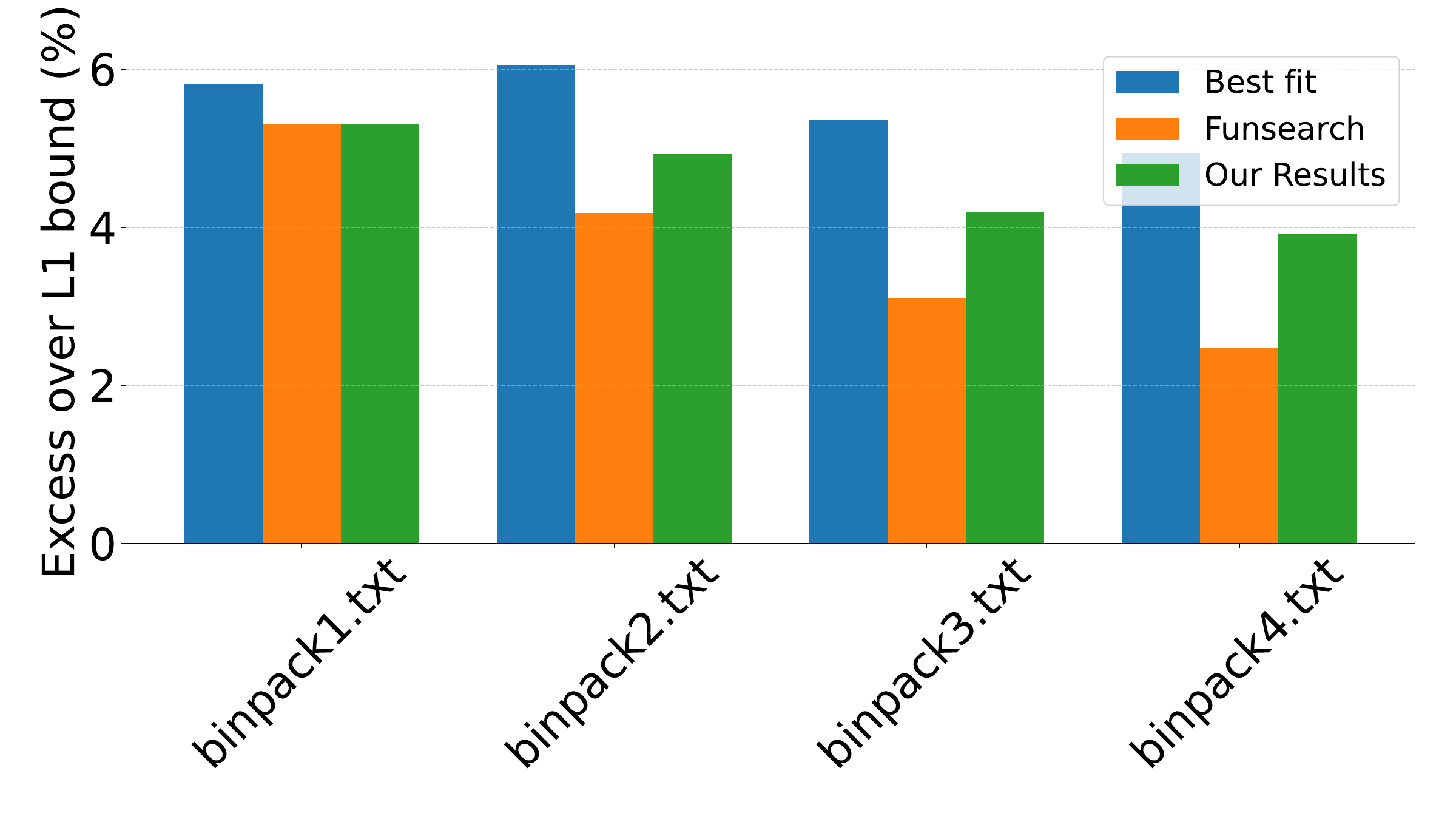}
    \caption{Comparisons of the accuracy of discovered algorithms for the bin packing task.}
    \label{fig:binpacking}
\end{figure}

We verified our LLM-assisted ES implementation by applying it to the bin-packing task as defined in the DeepMind FunSearch paper \cite{RomeraParedes2024}. By doing so, we are able to benchmark our implementation against FunSearch. We note that apart from potential differences in the implementation, we also rely on a LLM from DeepSeek \cite{guo2025deepseek} in contrast to a Gemini variant for FunSearch.\\
The experiments are based on standard benchmark datasets from the OR-Library \cite{beasley1990or}, including \texttt{binpack1}, \texttt{binpack2}, \texttt{binpack3}, and \texttt{binpack4}. Each dataset contains 20 bin packing instances with 120, 250, 500, and 1000 items, respectively. The sizes of the items are sampled uniformly from the interval $[20,100]$ and the bin capacity is fixed at 150. Following the original setup, we generated a training dataset of 20 instances with 120 items (mirroring \texttt{binpack1}) and a validation dataset of 20 instances with 250 items (similar to \texttt{binpack2}). Candidate heuristics were evolved using our LLM-assisted ES and selected based on validation performance. The best performing programs were then evaluated on the complete benchmark suite (\texttt{binpack1}–\texttt{binpack4}) to assess generalization. This reproduction served to confirm that our implementation faithfully replicates the original methodology and yields consistent performance behavior.\\
In Figure \ref{fig:binpacking} and Table \ref{table:binpackingtable}, we compare the performance of three heuristics on the OR-Library bin packing benchmarks: the standard Best Fit heuristic, the heuristic discovered by the original DeepMind FunSearch framework, and the one produced by our own implementation.\\
Across all datasets, both Deepmind FunSearch and our implementation outperform the default Best Fit heuristic, confirming that learned algorithms lead to more efficient bin usage. In \texttt{binpack1.txt}, both Deepmind FunSearch and our algorithm achieve a 5.30\% excess over the L1 bound, improving on the Best Fit heuristic's 5.81\%. In \texttt{binpack2.txt}, Deepmind FunSearch achieves the best result (4.19\%), with our method slightly behind at 4.92\%, both improving significantly over Best Fit heuristic's 6.06\%. A similar trend holds for \texttt{binpack3.txt}, where Deepmind FunSearch yields the lowest excess (3.11\%), followed by our result at 4.20\%, again outperforming the Best Fit heuristic (5.37\%). In \texttt{binpack4.txt}, the pattern continues: Deepmind FunSearch produces the best excess (2.47\%), our method is second (3.92\%), and Best Fit trails behind (4.94\%).
Regarding computational cost, our method was run for six days on four H100 GPUs.  The original DeepMind FunSearch paper reports the heuristic used for bin packing, but does not disclose the exact number of runs required to obtain it. Based on the scale of their capset experiments (which involved 140 separate runs, each using 15 A100 GPUs for two days), it is possible that their original search could have involved substantially more compute than our replication. Despite this, our results remain highly competitive.\\
These results demonstrate that our implementation is capable of discovering heuristics that outperform the current Best Fit heuristic and are competitive with those produced by DeepMind’s FunSearch framework.  

\begin{table}[ht]
\centering
\renewcommand{\arraystretch}{1.1}
\setlength{\tabcolsep}{8pt}
\begin{tabular}{lccc}
\toprule
\textbf{Dataset} & \textbf{FunSearch} & \textbf{Our Results} & \textbf{Best Fit} \\
\midrule
\texttt{binpack1} 
& \cellcolor{green!60}5.30\% 
& \cellcolor{green!60}5.30\% 
& \cellcolor{red!45}5.81\% \\

\texttt{binpack2} 
& \cellcolor{green!60}4.19\% 
& \cellcolor{yellow!25}4.92\% 
& \cellcolor{red!45}6.06\% \\

\texttt{binpack3} 
& \cellcolor{green!60}3.11\% 
& \cellcolor{yellow!25}4.20\% 
& \cellcolor{red!45}5.37\% \\

\texttt{binpack4} 
& \cellcolor{green!90}2.47\% 
& \cellcolor{yellow!25}3.92\% 
& \cellcolor{red!45}4.94\% \\
\bottomrule
\end{tabular}
\caption{Excess over L1 bound (\%) for each method on binpack datasets. 
Lower is better. Colors indicate performance per row: 
\textcolor{green!80!black}{green} = best, 
\textcolor{yellow!70!black}{yellow} = middle, 
\textcolor{red!80!black}{red} = worst.}
    \label{table:binpackingtable}

\end{table}

\newpage
\section{Appendix B: Hyperparameters}
\label{app:hyper}

In Table \ref{table:hyperparameters} we report the hyperparameters used for the LLM-assisted ES and CGP frameworks.

\begin{table}[ht]
\centering
\renewcommand{\arraystretch}{1.1}
\setlength{\tabcolsep}{5pt}
\begin{tabular}{|l|c|c|}
\hline
\textbf{Parameter} & \textbf{LLM-assisted ES} & \textbf{CGP} \\
\hline
\textbf{Database Size} 
& 200 
& 200 \\
\hline
\textbf{Sampling Temperature} 
& 0.2 
& 0.2 \\
\hline
\textbf{Prompts per Iteration} 
& 30 
& NA \\
\hline
\textbf{Tokens per prompt} 
& 3000
& NA \\
\hline
\textbf{Mutations per Iteration} 
& varies 
& 1000 \\
\hline
\textbf{Parents per Mutation} 
& 2 
& 1 \\
\hline
\textbf{Max Algorithm Size} 
& NA
& target size + 2\\
\hline
\end{tabular}
\caption{Comparison of hyperparameters used in LLM-assisted ES and CGP.}
\label{table:hyperparameters}
\end{table}

\section{Appendix C: Dynamical System}
\label{sec:environment}
We present the dynamical system which is used as an illustrative example within this paper in more detail:\\
We consider a linear time-invariant discrete-time dynamical system modeling the position and velocity of an object. The continuous-time second-order equation is discretized using a fixed sampling interval \( \Delta t \), yielding the following stochastic difference equation:
\vspace{-0.5em}
\begin{equation}
    \mathbf{x}_k = \mathbf{F} \mathbf{x}_{k-1} + \mathbf{G} a_k.
\end{equation}

Here, \( a_k \) is a random acceleration input modeled as a zero-mean Gaussian variable, and the state vector \( \mathbf{x}_k = [p_k,\, v_k]^T \) includes position and velocity. The system matrices are defined as:
\vspace{-0.5em}
\[
\mathbf{F} = \begin{bmatrix} 1 & \Delta t \\ 0 & 1 \end{bmatrix}, \quad
\mathbf{G} = \begin{bmatrix} \frac{1}{2} \Delta t^2 \\ \Delta t \end{bmatrix}.
\]

This corresponds to the forward-Euler discretization of Newtonian motion under stochastic acceleration. The acceleration noise \( a_k \sim \mathcal{N}(0, \sigma_a^2) \) induces process noise \( \mathbf{w}_k = \mathbf{G} a_k \), distributed as:
\vspace{-0.5em}
\[
\mathbf{w}_k \sim \mathcal{N}(\mathbf{0}, \mathbf{Q}), \quad \text{with} \quad
\mathbf{Q} = \mathbf{G} \mathbf{G}^\top \sigma_a^2 =
\sigma_a^2 \begin{bmatrix}
\frac{1}{4} \Delta t^4 & \frac{1}{2} \Delta t^3 \\
\frac{1}{2} \Delta t^3 & \Delta t^2
\end{bmatrix}.
\]

For simplicity, we assume full observability of the system state at each time step. Observations are obtained through a linear measurement model:
\vspace{-0.5em}
\[
\mathbf{z}_k = \mathbf{H} \mathbf{x}_k + \mathbf{v}_k, \quad
\mathbf{H} = \begin{bmatrix} 1 & 0 \\ 0 & 1 \end{bmatrix},
\]
where both position and velocity are measured directly. The measurement noise \( \mathbf{v}_k \) is modeled as zero-mean Gaussian with covariance \( \mathbf{R} = \sigma_z^2 \mathbf{I} \), i.e.,
\vspace{-0.5em}
\[
\mathbf{v}_k \sim \mathcal{N}(\mathbf{0}, \sigma_z^2 \mathbf{I}).
\]

The resulting discrete-time state-space model satisfies the standard conditions for applying the Kalman filter: linear time-invariant dynamics, additive Gaussian noise in both the process and measurements, and fully known system parameters and noise covariances. Under these assumptions, the Kalman filter yields the optimal state estimate in the minimum mean square error (MMSE) sense. Due to its analytical tractability, physical relevance, and ability to capture essential features of uncertainty propagation in dynamic systems, the described model is widely considered a canonical benchmark for evaluating state estimation methods in filtering, tracking, and control applications \cite{greenberg2023optimization,freirich2023perceptual}.

\section{Appendix D: Progressive Discovery}
\label{appendix:progressive-discovery}

In this section, we continue the experiment of rediscovering the Kalman-Filter by analyzing the differences between LLM-assisted ES and CGP in more detail. To do so, we run multiple experiments and task our framework with discovering a progressively more complete Kalman Filter algorithm.\\
We begin with the simplest case: the prediction step of the Kalman filter. The corresponding computational procedure is shown in the first two lines of Figure~\ref{fig:kalman_target_a}. We progressively increase the complexity of the discovery task by incrementally introducing additional components of the Kalman filter, adding more and more operations at a time to the target algorithm until the complete filter is recovered.\\ 
Table~\ref{tab:method_comparison_a} presents the MSE loss for the different approaches and experiments. We observe that both the \textit{predict}, \textit{predict + 12.5\% update}, and \textit{predict + 25\% update}  were successfully discovered by both CGP and LLM-assisted ES. In particular, CGP was also able to discover the more complex variants, while maintaining MSE values closely aligned with the optimal solution. This consistent accuracy across increasing program complexity highlights the robustness of CGP. However, LLM-assisted ES seems to get stuck in a local minima and only finds algorithms that produce a forecast with high MSE. This behavior is likely attributable to its reliance on a LLM comprising only 14 billion parameters. Moreover, even a LLM of this size must be executed on GPUs and incurs significant computational overhead. Consequently, LLM-assisted ES is limited in the number of tokens it can evaluate within a fixed resource budget. We hypothesize that a larger computational budget would increase the abilities of LLM-assisted ES, which is in accordance with new research \cite{novikov2025alphaevolve}.\\
The results also justify our choice of running both approaches, i.e. CGP and LLM-assisted ES in parallel, to facilitate robust algorithmic discovery.\\
We also observe a consistent trend in the performance of Random Search. While it can occasionally identify competitive solutions for simple tasks, its effectiveness deteriorates as program complexity increases. This is expected, as more complex problems yield a significantly larger solution space, reducing the probability of discovering optimal solutions through purely random exploration. 
\begin{table*}[ht!]
\centering
\setlength{\tabcolsep}{28pt}
\renewcommand{\arraystretch}{1.1}
\begin{tabular}{lccc}
\toprule
\textbf{Method} & \textbf{CGP} & \textbf{LLM-assisted ES} & \textbf{Random Search} \\
\midrule
\textbf{predict}               
& \cellcolor{green!60}$0.995077 \pm 9\mathrm{e}{-3}$ 
& \cellcolor{green!45}$0.995493  \pm 9\mathrm{e}{-3} $ 
& \cellcolor{yellow!30}$1.009967  \pm 9\mathrm{e}{-3}$  \\

\textbf{predict + 12.5\% update}      
& \cellcolor{green!60}$0.995077 \pm 9\mathrm{e}{-3}$  
& \cellcolor{green!45}$0.996792 \pm 9\mathrm{e}{-3}$  
& \cellcolor{orange!30}$1.069877 \pm 1\mathrm{e}{-2}$ \\

\textbf{predict + 25 \% update}     
& \cellcolor{green!60}$0.995077 \pm 9\mathrm{e}{-3} $ 
& \cellcolor{orange!30}$1.042292  \pm 9\mathrm{e}{-3}$ 
& \cellcolor{orange!30}$1.045160 \pm 8\mathrm{e}{-3}$  \\

\textbf{predict + 50 \% update}    
& \cellcolor{green!45}$0.994762 \pm 9\mathrm{e}{-3} $ 
& \cellcolor{red!25}$1.968045 \pm 1\mathrm{e}{-2} $ 
& \cellcolor{red!15}$1.141732 \pm 1\mathrm{e}{-2}$  \\

\textbf{predict + 75 \% update}     
& \cellcolor{green!45}$1.003259 \pm 9\mathrm{e}{-3} $ 
& \cellcolor{red!25}$1.968045  \pm 1\mathrm{e}{-2} $ 
& \cellcolor{red!15}$1.209344  \pm 1\mathrm{e}{-2}$ \\

\textbf{predict + update}      
& \cellcolor{green!45}$0.995220 \pm 9\mathrm{e}{-3} $ 
& \cellcolor{red!25}$1.968045  \pm 1\mathrm{e}{-2}$  
& \cellcolor{red!15}$1.209344  \pm 1\mathrm{e}{-2}$  \\
\bottomrule
\end{tabular}
\caption{MSE loss of our methods compared to Kalman Filter performance (0.995077). 
\textcolor{green!60!black}{Green} highlights optimal or near-optimal results, 
\textcolor{orange!80!black}{yellow/orange} indicates moderate deviation, and 
\textcolor{red!80!black}{red} marks the least favorable outcomes in each row.}
\label{tab:method_comparison_a}
\vspace{-1.em}
\end{table*}

\begin{figure}[ht]
\centering
\begin{pythonbox}
\begin{lstlisting}[style=pythonstyle]
def kalman(x, F, P, Q, z, R):
    # Predict step
    x_predict = F @ x
    P = F @ P @ F.T + Q
    # Update step
    y = z - x_predict
    S = P + R
    K = P @ inv(S)
    x_update = x_predict + K @ y
    P = P - K @ P
    return x_predict, P, y, S, K, x_update
\end{lstlisting}
\end{pythonbox}
\caption{{Kalman Filter algorithm}}
\label{fig:kalman_target_a}
\end{figure}

\section{Appendix E: Discovered algorithms}
\label{app:algo}

In Figure~\ref{fig:discovered_functions_delayed_obs}, we present the algorithms discovered by FunSearch and CGP for the delayed observation scenario, while Figure~\ref{fig:discovered_functions_non_linear} shows the corresponding solutions for the case of nonlinear state transitions. These examples highlight the structural differences between the two approaches. LLM-assisted ES tends to produce programs with greater algorithmic richness and mathematical depth, making use of a wide range of operations, including matrix products, element-wise nonlinearities, and adaptive scaling mechanisms. In contrast, the CGP-generated functions are more constrained in form and typically rely on a pre-defined, fixed-size set of operations.

One of the key strengths of LLM-assisted ES is its ability to discover expressive and complex update rules that go beyond the structure of classical filtering techniques. For instance, the inclusion of operations such as \texttt{np.log}, \texttt{np.tanh}, and sinusoidal terms in LLM-assistes ES outputs reflects an inherent flexibility to model richer dynamics or nonlinear observation processes. Moreover, in our case, these algorithms often maintain recognizable algorithmic blocks — such as state predictions \( xp \), covariance updates \( P \), and innovation terms \( y \) — which can be directly associated with standard Kalman filter notation. This facilitates theoretical analysis and easily shows in which part the newly discovered algorithm differs from the established baseline.

\begin{figure*}[t]
\centering
\noindent
\begin{minipage}[t]{0.49\textwidth}
  \begin{pythonbox}[height=5.5cm]
  \begin{lstlisting}[style=pythonstyle]
def function_approximate(x, F, P, Q, z, R):
    a = F @ x
    b = F @ np.log(np.maximum(a * 0.03, 1e-8))
    c = F @ np.tanh(b * 0.2)
    xp = a + c
    P = F @ P @ F.T + 0.7 * Q
    y = z - xp
    S = P + 0.7 * R
    inv_S = np.linalg.inv(S)
    K = P @ inv_S + 0.15 * inv_S
    x = xp + K @ y
    x += 0.6 * F @ np.tanh(F @ x * 0.08)
    P = (np.eye(F.shape[0]) - K) @ P
    return xp, P, y, S, K, x

  \end{lstlisting}
  \end{pythonbox}
\end{minipage}%
\hfill
\begin{minipage}[t]{0.49\textwidth}
  \begin{pythonbox}[height=5.5cm]
  \begin{lstlisting}[style=pythonstyle]
def graph_approximate(x, F, P, Q, z, R):
    A = F @ R
    B = P + Q
    K = np.linalg.inv(B)
    S = Q @ K
    y = A - x
    T = K.T
    P = S - S
    F_ = S + F
    x = y @ T + x
    S = F_ @ S
    K = T - P
    return xp, P, y, S, K, x
  \end{lstlisting}
  \end{pythonbox}
\end{minipage}
\caption{{ Algorithms discovered for delayed observations by LLM-assisted ES on the left and CGP on the right.}}
\label{fig:discovered_functions_delayed_obs}

\end{figure*}
\begin{figure*}[t]
\centering
\noindent
\begin{minipage}[t]{0.49\textwidth}
  \begin{pythonbox}[height=9.cm]
  \begin{lstlisting}[style=pythonstyle]
def function_approximate(x, F, P, Q, z, R):
    x = np.array(
        [0.04 * x[0]**3 - 1.8 * x[0] + 0.34 * np.sin(x[1]),
                  0.14 * np.tanh(0.05 * x[0] * (x[1] + 0.8))])
    xp = F.dot(x)
    scale_Q = (x[0] * x[1] + x[0]**2 + x[1]**2 + 0.6) * \
                (1 + 0.9 * (x[0] * x[1] + x[0]**2 + x[1]**2))
    P = F.dot(P.dot(F.T)) + Q * scale_Q
    y = z - xp
    scale_R = (x[0]**2 + x[1]**2 + 0.5) * (1 + 0.8 * (x[0]**2 + x[1]**2))
    S = P + R * scale_R
    inv_S = np.linalg.inv(S + 0.0002 * np.eye(S.shape[0]))
    K = P.dot(inv_S) * (0.85 * np.tanh(np.linalg.norm(y)))
    x = xp + K.dot(y)
    P = (np.eye(F.shape[0]) - K) * P * (0.5 + 0.05 * np.mean(y**2))
    return xp, P, y, S, K, x
  \end{lstlisting}
  \end{pythonbox}
\end{minipage}%
\hfill
\begin{minipage}[t]{0.49\textwidth}
  \begin{pythonbox}[height=9.cm]
  \begin{lstlisting}[style=pythonstyle]
def graph_approximate(x, F, P, Q, z, R):
    x = np.array([0.05 * x[0]**3 - 2 * x[0], 0.1 * np.sin(x[1])])
    Ft = F.T
    y = z - (F @ x)
    S = Ft.T + Ft + P + Q
    K = Ft @ np.linalg.inv(S)
    P = K @ y
    x = x + P
    return xp, P, y, S, K, x
  \end{lstlisting}
  \end{pythonbox}
\end{minipage}
\caption{{ Algorithms discovered for non linear transitions by LLM-assisted ES on the left and CGP on the right.}}
\label{fig:discovered_functions_non_linear}
\end{figure*}

\end{document}